\documentclass[sigconf]{acmart}
\AtBeginDocument{%
  \providecommand\BibTeX{{%
    \normalfont B\kern-0.5em{\scshape i\kern-0.25em b}\kern-0.8em\TeX}}}

\setcopyright{acmcopyright}
\copyrightyear{2022}
\acmYear{2022}
\acmDOI{XXXXXXX.XXXXXXX}

\acmConference[]{}{}{}
\acmSubmissionID{tcom_101}

\usepackage{algorithm}
\usepackage{algpseudocode}
\begin{document}

\title{Zero-Shot Multi-Modal Artist-Controlled Retrieval and Exploration of 3D Object Sets}
%\title{Zero-Shot Multi-Modal Sketch Informed Retrieval and Exploration of 3D Object Sets}
%\title{SEARCH: Semantically Embedded Artistic Retrieval Control in Hyper-dimensional space}
%\title{Z-SEARCH:  Zero-shot Semantic Embeddings for Artistic Retrieval and Control in Hyper-dimensional space}


\author{Kristofer Schlachter}
\authornote{Both authors contributed equally to this research.}
\email{kristofer@unity3d.com}
\orcid{0000-0001-6735-5867}
\affiliation{%
  \institution{Unity Technologies}
  %\city{San Francisco}\state{California}
  \country{U.S.A.}
}

\author{Benjamin Ahlbrand}
\authornotemark[1]
\email{ben.ahlbrand@unity3d.com}
\orcid{0000-0002-7616-9625}
\affiliation{
  \institution{Unity Technologies}
  %\city{San Francisco}\state{California}
  \country{U.S.A.}
}

\author{Zhu Wang}
\email{zhu.wang@nyu.edu}
\orcid{0000-0003-2936-5479}
\affiliation{
  \institution{New York University}
  %\city{New York}\state{NY}
  \country{U.S.A} 
}

\author{Valerio Ortenzi}
\email{valerio.ortenzi@unity3d.com}
\orcid{0000-0002-8194-1616}
\affiliation{
  \institution{Unity Technologies}
  %\city{Copenhagen}
  \country{Denmark}
}

\author{Ken Perlin}
\email{perlin@cs.nyu.edu}
\affiliation{
  \institution{New York University}
  %\city{New York}\state{NY}
  \country{U.S.A} 
}

\renewcommand{\shortauthors}{Schlachter and Ahlbrand, et al.}

\begin{abstract}
When creating 3D content, highly specialized skills are generally needed to design and generate models of objects and other assets by hand. We address this problem through high-quality 3D asset retrieval from multi-modal inputs, including 2D sketches, images and text. We use CLIP as it provides a bridge to higher-level latent features. We use these features to perform a multi-modality fusion to address the lack of artistic control that affects common data-driven approaches. Our approach allows for multi-modal conditional feature-driven retrieval through a 3D asset database, by utilizing a combination of input latent embeddings. We explore the effects of different combinations of feature embeddings across different input types and weighting methods.
\end{abstract}




\begin{teaserfigure}
\centering
  \includegraphics[width=\textwidth]{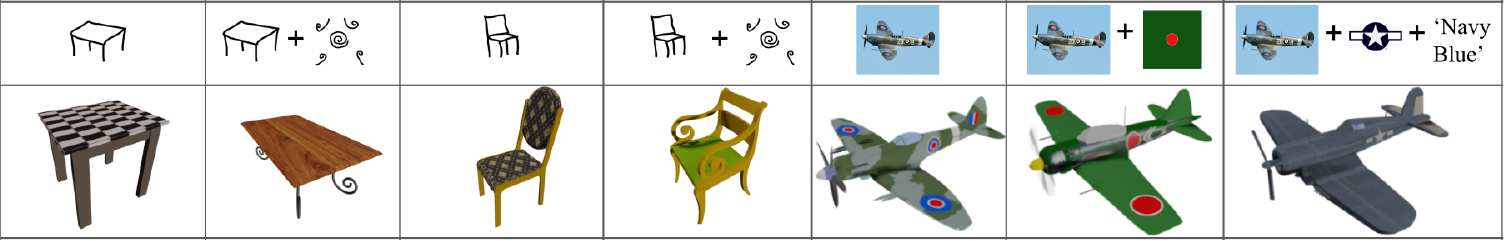}
  \caption{Examples of retrieval with only the top match shown.
%   \medskip
%   \small
  Top Row: \textmd{Queries containing sketch, image, and text inputs to the 3D mesh retrieval system  (weights are not shown in the multiple input queries).} Bottom Row: \textmd{Top match in the 3D mesh database based on the similarity score of the query's embedding with the embedding of the renderings of the 3D meshes.}}
  \label{fig:teaser}
\end{teaserfigure}

\maketitle
%!TEX root = main.tex
\section{Introduction}

The creation of 3D content generally presents a high barrier to entry, since it requires specialised skill sets even when using existing modelling software. While artists and other experienced content creators can navigate their way through labor-intensive pipelines, other less skilled users are often discouraged and tend to give up. For example, hobbyist world builders are generally able to produce a rough sketch of a 3D model, but then often find the process of creating the corresponding 3D content to be long and difficult. Furthermore, we observe a disconnect between creative workflows and existing sketch-to-mesh retrieval-based methods, as such methods tend to lack the artistic control needed to explore the space of results.

In this paper, we address the problem of democratising content creation by enabling 3D asset retrieval from 2D images, simple 2D sketches, and text input. In particular, our approach allows users to define the weights of inputs by leveraging CLIP (Contrastive Language-Image Pre-Training) embeddings \cite{radford2021learning}. We strongly believe that a system which accepts diverse inputs such as sketches, 2D images and text can further lower the barrier to entry for 3D content creation. From this perspective, the main contributions of this paper are:

\begin{itemize}
    \item A novel pipeline with zero-shot sketch-based multi-modal retrieval.
    \item A weighted interpolation in the latent space of (visual style independent) multi-modal inputs to enable artistic control over the retrieval sets.
\end{itemize}
%!TEX root = main.tex
\section{Related Work}
Our work builds upon recent CLIP-like models. CLIP is a neural network optimized over 400 million natural images and matching text descriptions. Its embedding space allows ``semantic alignment'' of features between images and text passed as input. Our proposed method enables a feature extraction which allows manipulation of the latent space such that desirable semantic traits can be retrieved without the need for a complex dataset or training process.

% consider adding
% SkexGen learns decompositions of sketches into multiple codebooks...https://samxuxiang.github.io/skexgen/index.html  \cite{https://doi.org/10.48550/arxiv.2207.04632}

%CLIP (Contrastive Language-Image Pre-training) is a neural network that connects image and text, allows image prediction from text supervision.
%clip-matrix \cite{https://doi.org/10.48550/arxiv.2109.12922} is a simple direct optimization of a SMPL mesh parameters using a differentiable renderer and CLIP text prompts as supervision.  They also introduce combining multiple text encodings by summing them.
    
\subsection{CLIP-driven Generative Models} Other works use CLIP-encoded 2D images and text to accomplish tasks involving 3D meshes. Text2Mesh\cite{michel2021text2mesh} and Text-to-Mesh \cite{khalid2022text} optimize 3D deformations of 3D meshes by combining a differentiable renderer with supervisory CLIP text prompts. Text2Mesh also explored optimizing towards multiple input targets by summing the embeddings. Our methods allow for arbitrary weightings of embeddings and real-time feedback of such effects.

We use 2D CLIP encoded views of a 3D object in a similar manner to Dreamfields\cite{jain2021dreamfields}. However, in \cite{jain2021dreamfields} the similarity of the input text prompt supervises the NeRF's optimization across multiple camera views while our approach uses these prompt features to compute similarity for retrieval.

%CLIP has also been used to allow text based synthesis of a NeRF (Neural Radiance Field). In Dreamfields\cite{jain2021dreamfields} the similarity of the input text prompt supervises the NeRF's optimization across multiple camera views.

%CLIPASSO \cite{vinker2022clipasso}  uses CLIP embeddings to ensure that more and more abstract sketches of the same object still are semantically the same by using the penalizing drift away from an embedding of a text prompt describing the family of sketches.
%Text2Mesh\cite{michel2021text2mesh} is a recent work that uses CLIP embedding based optimization of a Neural Style Field (NSF) that colors a high quality input mesh and does local deformation.  It imposes both a neural and geometric prior and allows for multiple modes of stylization along with detail control through a enhanced positional encoding of input vertex positions.

Our work uses CLIP to semantically understand drawings. This was also done in CLIPASSO \cite{vinker2022clipasso} where the authors used it to ensure increasingly abstract sketches that are semantically similar to the supervisory text prompt.
Concurrent work to ours called TASK-former \cite{tsbir2022} sums the embedding of a single sketch and a single text prompt to retrieve images. That work shows how using a text prompt embedded by CLIP can achieve state-of-the-art results via text-based image retrieval. Our work differs by allowing any number of inputs and modes (in fact, text is not required), and allowing custom weights on those inputs; furthermore, we do not need to train on a dataset, due to our zero-shot approach.

%Text-to-Mesh\cite{khalid2022text} optimizes subdivision surfaces using a differentiable renderer and CLIP embedding of text description. CLIP's joint text and image embeddings allow for a similarity based loss to be used between the text input and the rendering. The subdivision surface is directly optimized which leads to smooth geometric prior and speed-up in training. 

%Dreamfields\cite{jain2021dreamfields} optimizes an MLP based neural radiance field using a differentiable renderer and CLIP embedding of text description. CLIP's joint text and image embeddings allow for a similarity based loss to be used between the text input and the rendering. They use a transmittance prior to reduce artifacts in synthesized radiance fields. 

% still relevant to the current work?
% VolSDF \cite{yariv2021volume} improves geometry reconstruction from a Neural Radiance field by representing density as the CDF of a Laplace distribution of an SDF.  It succeeds in separating/disentangling the geometric surface from the radiance. 

% is this relevant?
% clip-forge \cite{sanghi2021clip} voxels

\subsection{Zero-Shot Sketch-based Retrieval}
% \cite{pu20053d} must we cite classic work that predates either above sections in a short paper?
There are a number of recent works on zero-shot sketch-based image retrieval.
Domain Disentangled GANs (DD-GAN) \cite{xu2022domain} train a generative adversarial network on SHREC '13 \cite{li2013shrec} + SHREC '14 \cite{li2014shrec} in order to build a mapping between user-based sketches and the 3D model to be retrieved. In practice, this is a brittle approach, since a wide range of inputs can be objectively ``semantically correct'' in representing a shape. Su et al. \cite{su2022semantically} built a bidirectional projection between multi-view information of 3D shapes and semantic attributes. In contrast, we use a proven pre-trained model with existing features for semantic alignment in order to alleviate dataset bias\slash constraints. This allows us to handle a wide band of possibilities.

Tursun et. al. \cite{efficient_zs_sbir} use a pre-trained model as a teacher to help learn abstract labels for their zero-shot sketch-based image retrieval (SBIR) model. Dutta and Akata\cite{cycle_zs_sbir} use an adversarial loss and enforce cycle-consistency on the shared latent space to close the domain gap. In comparison, CLIP uses contrastive learning and text descriptions of 29 times more images than ImageNet \cite{imagenet} to create a shared latent space.  

To our knowledge, all existing sketch-based retrieval methods (including zero-shot) return a one-to-one mapping between sketch and result. They are further limited by a single sketch style and by a lack of multi-modal input, as well as by a single representation (style, level of abstraction, etc).

%confirm citation is correct
Previous works\cite{michel2021text2mesh} have also shown that 2D viewpoint supervision has been effective for 3D tasks. Those same works have emphasized that changing the viewpoint of the object doesn't change its semantics. In contrast, our work allows the user to distort the mapping through a combination of many inputs and\slash or weighting of those inputs in order to explore the output space. In this work, we explore the effects of different combinations of inputs across different media types and weighting methods.

\section{Method}

\subsection{Zero-shot learning using CLIP}
Zero-shot learning methods allow for the classification\slash interpretation of data that was not seen during training.  Zero-Shot models produce a rich embedding space that aims to optimize identification of semantically meaningful features, which can then be used to cluster around desired categorizations. This is done through auxiliary information, which in CLIP's case is textual descriptions of images. The learned semantic features are useful for down-stream tasks such as retrieval where unseen images can be embedded into the rich feature space and a distance measure can be performed.
% some of this might belong in related works (ie CLIP specific insights)
We used a pre-trained CLIP model\cite{radford2021learning} because it was optimized on the task of ``representation learning''. CLIP also has the added benefit of aligning the embedding of multi-modal inputs (in this case images and text) into a common feature-rich embedding space that can be used for zero-shot based retrieval.  

\subsection{Data Setup}
%The retrieval database construction is accomplished by taking the CLIP encoders for text and images and using them to generate a latent vector for each of the query's inputs in either modality. Other input forms such as sketches are treated as images and 
Retrieval of 3D models is enabled by rendering each model from three different viewpoints (front, back and perspective) and storing the corresponding images along with each mesh. Then each image is encoded through CLIP and that embedding is stored along with the model. We found that these three views lead to valid matches for the specific models that were in the dataset.

The dataset itself is 5 categories of ShapeNet\cite{shapenet2015}: sofa, chair, table, car and plane (roughly 25K meshes). We embedded them using three different rendering setups: the original ShapeNet 3D mesh, the original model without textures, and finally a decimated, smoothed and untextured mesh. The untextured configurations helped explore the effect of texture and detail on matching for retrieval.

\subsection{Multi-Modal Embedding Fusion}
\begin{figure}
\includegraphics[width=0.9\columnwidth]{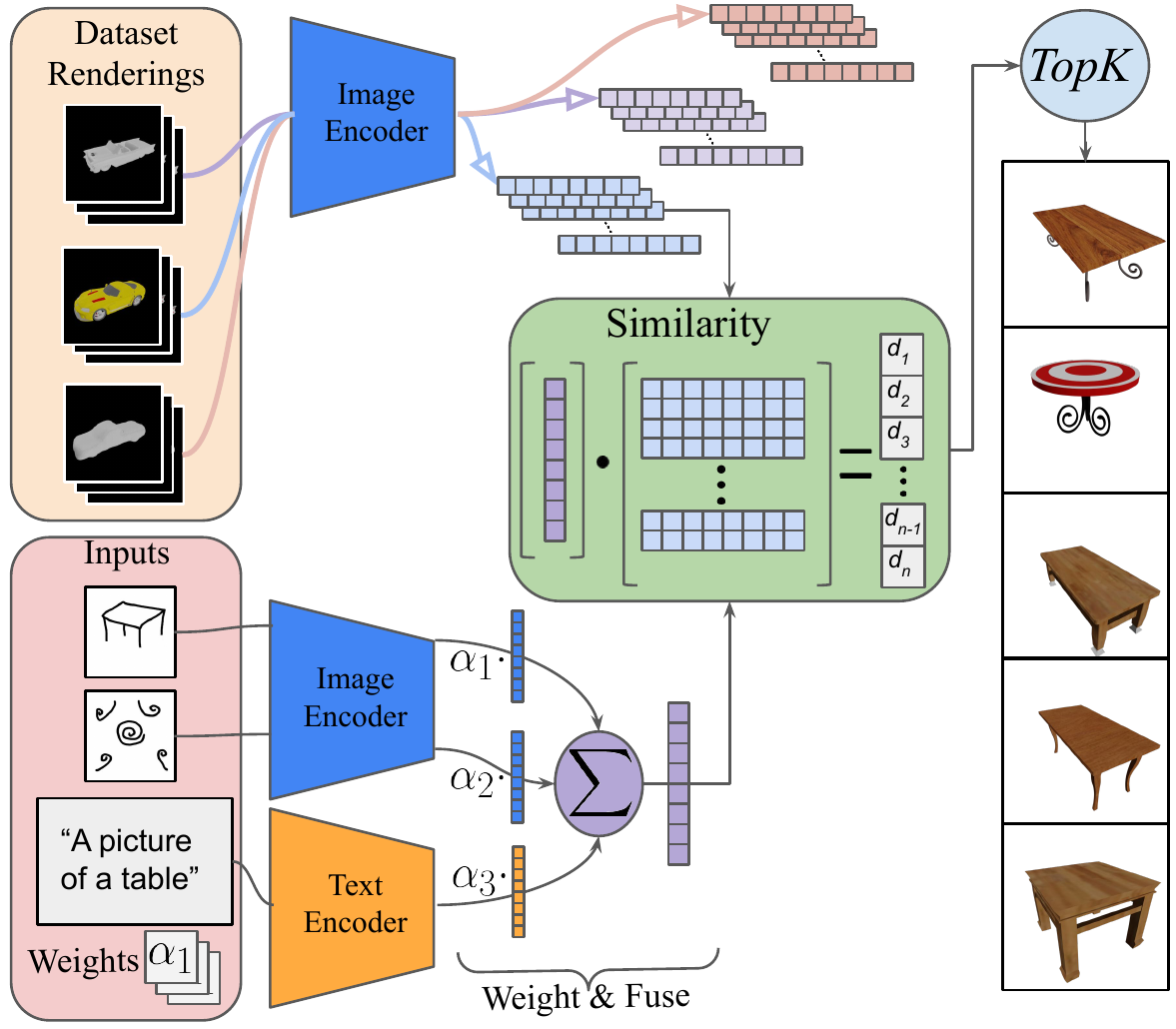}
\caption{Architectural overview of our method. \textmd{
The dataset is rendered three ways (textured, untextured, untextured and smoothed). The images are then embedded using CLIP's image encoder.  Each query input is separately CLIP-encoded, weighted and then fused into a single embedding. The fused query is then used to score each rendering in the active dataset (in this example textured) and the top 5 similarities per mesh are retrieved.
}
}
\label{fig:architecture}
\vspace{-4mm}
\end{figure}

These 2D renderings allow retrieval on 3D models to be effective. Since CLIP has aligned the embedding between text and images, queries for retrieval can take the form of text prompts and images. Sketches are passed through the CLIP image encoder and can be combined with an arbitrary number of other text or image inputs. Each is encoded and then combined into a single 512 dimension normalized code. After computing the cosine similarity score (see equation \ref{query-formula}) between that code and every 3D mesh's rendered images, we sort and filter (thereby removing duplicate matches that may result from multiple views being in the database) the top-k nearest neighbors. See figure \ref{fig:architecture} for more details on the data flow.

Embedding code fusion takes place in prior work \cite{michel2021text2mesh} through summing and averaging. We demonstrate this alongside arbitrary weighting for the embeddings. In order to arbitrarily weight inputs, we expose a single float per input that we then multiply by before summing (as opposed to simply summing unmodified embeddings).

%(EMPHASIZE IMPORTANCE OF ARTISTIC CONTROL OF ML OUTPUT) 
This weighting enables artistic control, in order to allow interactive refinement of the query embedding and to direct the user's search as a natural process. We show our weighted interpolation of multiple embeddings in equation \ref{weight-formula}, where $n$ refers to the number of inputs, $z$ refers to the embedding, and $\alpha$ is the user specified weights.
% \begin{equation} \label{average-formula}
%     fusion = \Sigma^{n}_{i=1} (1 / n) z \tag{average}
% \end{equation}
% \begin{equation} \label{sum-formula}
%     fusion = \Sigma^{n}_{i=1} z \tag{sum}
% \end{equation}

\begin{equation} \label{weight-formula}
    fusion = \Sigma^{n}_{i=1} \alpha z%\tag{weighted}
\end{equation}

\begin{equation} \label{query-formula}
    similarity = [query] [features]^T
\end{equation}
% Create Figure Of Architecture / Overview.
% Show that architecture makes it easy to switch matching across rendering and mesh fidelities. It also allows matching against all 3 mesh display modes.

% SEGUE INTO RESULTS!!!!!

\section{Results And Discussion}

To illustrate the flexibility of our method, we explore different image inputs, as well as the effect of texturing on retrieval queries containing high frequency class independent features, in addition to enabling fine-tuning different weighted combinations to arrive at a result.

\subsection{Robust Semantic Encoding of Images}
To demonstrate the robustness of matching through the style-invariant semantic embedding of input sketches, we chose a variety of artistic visual styles that were distinct from the style of the rendered objects in the dataset. Figure \ref{fig:styles} shows that a varied range of image fidelity, drawing methods and styles can produce valid matches. Retrieval is also able to handle incomplete renditions of objects, as in the chair and truck queries. Semantic understanding of the query is also not impacted when multiple objects are in the image as is the case for the couch and airplanes.
This may be due to the use of a zero-shot learner where features are learned over category labels with high level features to allow a correct match at various levels of abstraction.

\subsection{Fine Details Inform Retrieval}
\begin{figure} 
    \includegraphics[width=0.9\columnwidth]{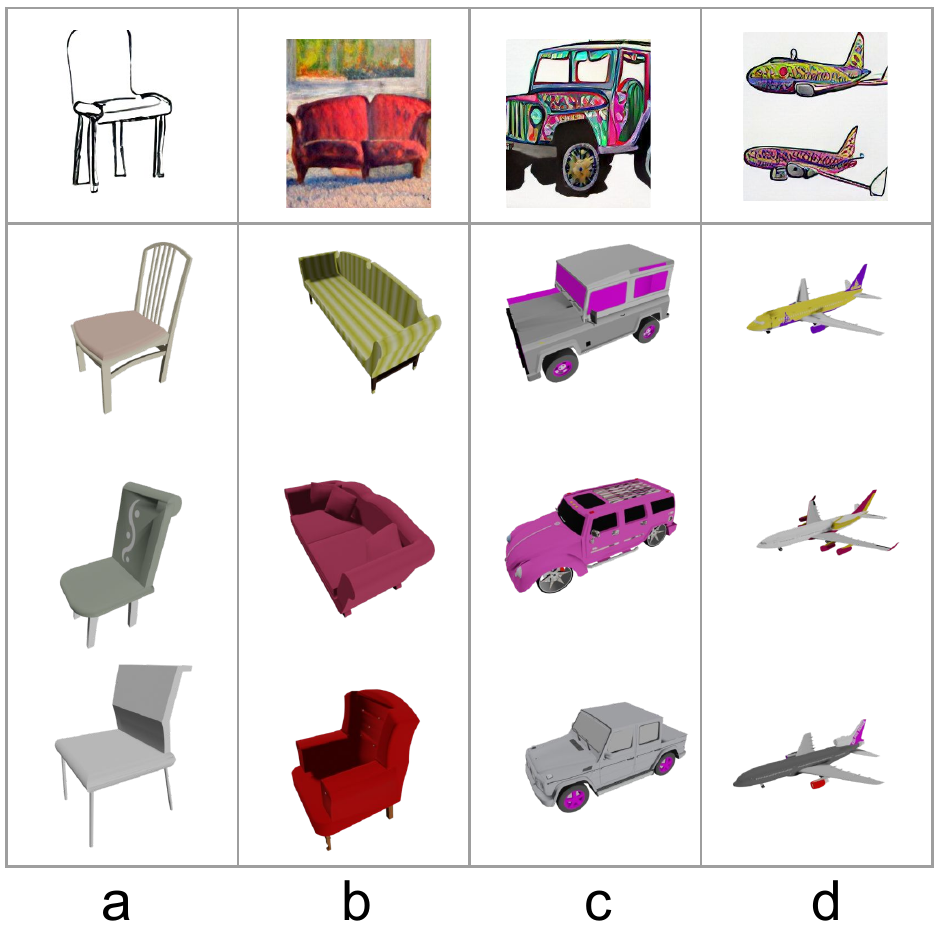}
    \setlength\abovecaptionskip{0.5\baselineskip}
    \caption{Examples of different input sketch styles and details. Top Row: \textmd{Inputs representing different level of abstraction and style.} Other Rows: \textmd{The top matches to the input query where the closest match is the top row and the next ranking ones are below.}}
    \label{fig:styles}
    \vspace{-5mm}
\end{figure}

Figure \ref{fig:styles} also shows that more detailed sketches can lead to more specific matches. Adding color, texture and details influence which model is retrieved. These cues, when available, disambiguate objects that belong to the same semantic category. This can be seen in the matching of the pink texture of the truck and more interestingly the results of the two airplane query matches airplanes that incorporate coloring of both.  A shortcoming of the color matching is that the top scoring couch is actually the color of the background. It is the second ranking match which has the color of the couch in the query image. 

\subsection{Multi-Modal Input and Refinement}
As we can see, steering a high level semantic match through a single sketch can be done by altering the strokes or by adjusting color and texture.  However, this way of querying requires more artistic skill to accurately express one's intention. Instead, using a different search paradigm that combines multiple targets into the query allows for expressing a search concept without requiring the skill to create an accurate visual representation. By leveraging the learned semantic features of the CLIP encodings and fusing multiple inputs that are feature specific instead of object class specific, an intuitive form of semantic based retrieval can be explored.  Figure \ref{fig:teaser} shows how this can be accomplished by providing a simple high level sketch and auxiliary examples of stylistic flourishes to furniture, insignias on an aircraft, or text describing paint color. %Including more than one mode of input has been shown [CITATION of sketch and text SBIR paper here] to be able to compensate for missing details in a sketch.

\begin{figure}[ht]
\centering
    \includegraphics[width=0.9\columnwidth]{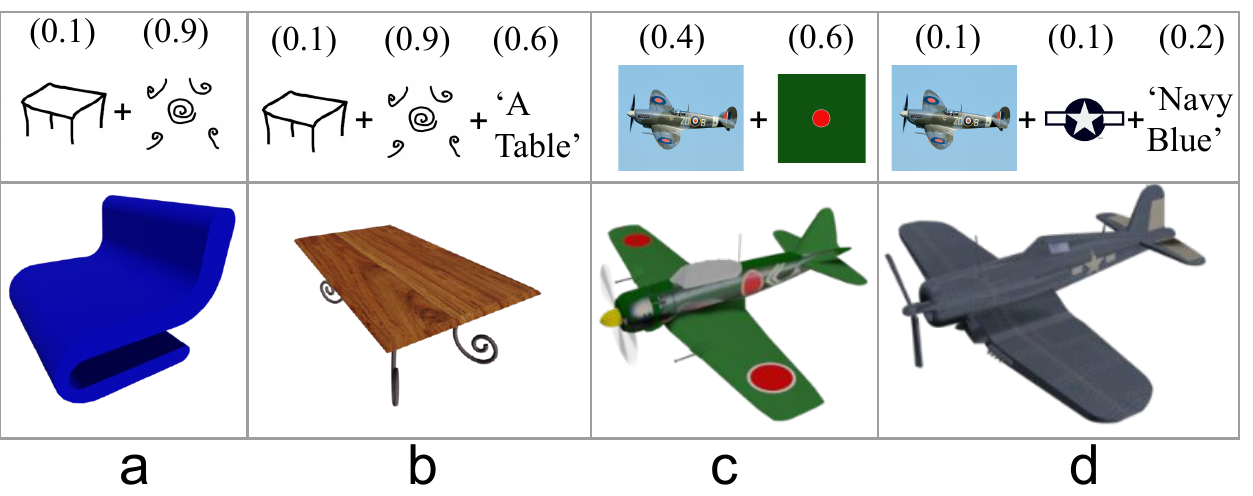}
    \setlength\abovecaptionskip{0.5\baselineskip}

    \caption{Examples weights shown for sub-feature matching. \textbf{Top Row}: \textmd{Inputs. The weights of the inputs are in parentheses and are not required to sum to one.} Bottom Row: \textmd{Top matching result. In column `a', notice the failure to match a table; adding a text prompt to the query improved matching. Columns `a' and `b' are matched against the untextured dataset to enforce geometric feature matching. The columns `c' and `d' were matched against the original textured models because the examples included texture details.}}
    \label{fig:weights}
    \vspace{-5mm}
\end{figure}

\subsection{Limitations and Steering The Result Set}
%The learned embedding has limitations and biases which may not interpret the scale or semantics of the query in the way you intended. 
Figure \ref{fig:weights}(a) shows that the query's encoding matches a chair better than a table.  It is also clear that a large flat surface and curves have been recognized and used for the match but are combined in an unintended way.  A solution is to reduce the space of possible matches: Figure \ref{fig:weights}(b) incorporates a text prompt that only encodes the features of a table and compels matches to be from the set of tables in the dataset.  

Other failure cases show up as interpreting the star insignia not as a texture but as geometry and matching furniture views that have the same rough outline. The context of the features also matters, just a red circle as an auxiliary input did not succeed in finding an airplane with that insignia, scaling it down and adding a green background resulted in the match shown in Figure \ref{fig:weights}(c). 
There is also a bias in the matching towards a subset of meshes if no auxiliary inputs are given. Conditioning on other inputs compensates for the biases of the encodings and allows wider exploration of the set of meshes. Some of the linear combinations of weights may not return intended results. Crucially, instant feedback to weight changes (less than 10 milliseconds on Titan RTX to query set of more than 100K embeddings) allows exploration of the fusion-space in real-time and thereby provides a practical way to interactively refine the result set to fit with one's artistic intentions.
\section{Future Work}

Future work should include exploring other text + image multi-modal models aside from CLIP such as DALLE-2 \cite{ramesh2022hierarchical}, and Imagen \cite{saharia2022photorealistic}, as well as fine tuning such models to provide better results on zero-shot sketch retrieval tasks.
We leave benchmarking as future work as we were unable to find any that necessarily included multi-modal inputs for zero-shot retrieval.
Our contribution to combining latent features is also applicable to other use cases such as image generation.

\section{Conclusion}

%This approach relies on the following assumptions:
%Our approach has the following shortcomings:
In this paper we addressed the problem of high-quality 3D asset
retrieval from multi-modal inputs, including 2D sketches, images and text.
We used CLIP features to perform multi-modality fusion in order to address the lack of artistic control that affects common data-driven approaches.
By aligning text and images we were able to perform retrieval at high levels of abstraction.
By using embeddings of 2D views of 3D meshes for retrieval we successfully captured view independent properties and semantics of the mesh, and allowed for effective retrieval. Finally, our weighted multi-input fusion created a pathway to help mitigate the biases and the idiosyncrasies of CLIP's embedded space.

% The examples.tex contains latex examples from the template
% \input{examples} 

\bibliographystyle{ACM-Reference-Format}
\bibliography{references}

% \appendix


% \input{sections/todo}
\end{document}